\documentclass[letterpaper]{article} 
\usepackage[submission]{aaai25}  
\makeatletter
\gdef\showauthors@on{T}
\makeatother
\usepackage{times}  
\usepackage{url}
\usepackage{cite}
\usepackage{multicol}
\usepackage{multirow}
\usepackage{helvet}  
\usepackage{courier}  
\usepackage{url}  
\usepackage{graphicx} 
\usepackage{natbib}  
\usepackage{caption} 
\usepackage{amsmath}  
\usepackage{appendix} 
\usepackage{algorithm}
\usepackage{algorithmic}

\usepackage{newfloat}
\usepackage{listings}
\DeclareCaptionStyle{ruled}{labelfont=normalfont,labelsep=colon,strut=off} 
\floatstyle{ruled}
\newfloat{listing}{tb}{lst}{}
\floatname{listing}{Listing}
\title{The Budget AI Researcher and the Power of RAG Chains}

\author {
    Franklin Lee\textsuperscript{\rm 1},
    Tengfei Ma\textsuperscript{\rm 2}
}
\affiliations {
    \textsuperscript{\rm 1}Jericho Senior High School, Jericho, NY\\
    \textsuperscript{\rm 2}SUNY Stony Brook University\\
    franklin.lee@jerichoapps.org, tengfei.ma@stonybrookmedicine.edu
}

\usepackage{bibentry}

\begin{document}

\maketitle

\begin{abstract}
Navigating the vast and rapidly growing body of scientific literature is a formidable challenge for aspiring researchers. Current approaches to supporting research idea generation often rely on generic large language models (LLMs). While LLMs are effective at aiding comprehension and summarization, they often fall short in guiding users toward practical research ideas due to their limitations. In this study, we present a novel structural framework for research ideation. Our framework, The Budget AI Researcher, uses retrieval-augmented generation (RAG) chains, vector databases, and topic-guided pairing to recombine concepts from hundreds of machine learning papers. The system ingests papers from nine major AI conferences, which collectively span the vast subfields of machine learning, and organizes them into a hierarchical topic tree. It uses the tree to identify distant topic pairs, generate novel research abstracts, and refine them through iterative self-evaluation against relevant literature and peer reviews, generating and refining abstracts that are both grounded in real-world research and demonstrably interesting. Experiments using LLM-based metrics indicate that our method significantly improves the concreteness of generated research ideas relative to standard prompting approaches. Human evaluations further demonstrate a substantial enhancement in the perceived interestingness of the outputs. By bridging the gap between academic data and creative generation, the Budget AI Researcher offers a practical, free tool for accelerating scientific discovery and lowering the barrier for aspiring researchers. Beyond research ideation, this approach inspires solutions to the broader challenge of generating personalized, context-aware outputs grounded in evolving real-world knowledge.\end{abstract}

Code — 

\url{https://github.com/hellojoeAoPS11235/ai-research-agent/tree/main}

\section{Introduction}
One  of the earliest uses of AI in research is in data analysis. Before the rise of ChatGPT and other advanced AI tools, researchers hoped that AI would help scientific researchers analyze data, leading to discoveries regarding variables' relationships \cite{8}. AI has come a long way since then: it can help scientists read papers and analyze them for potential hypotheses.

Getting a shallow understanding of a research paper usually takes up to 1 hour, and getting a deep understanding usually takes up to 6, depending on the field of study \cite{11}. About 23$\%$ of a scientist’s work time is spent reading papers over all career stages from undergraduate students to professors (academics), and it can be frustrating, especially for more inexperienced researchers \cite{12}.

With the rise of artificial intelligence in recent years, researchers developed numerous models every year. In 2023 alone, significant AI advances like GPT-4, Gemini, and DALL-E 3 opened to the public. With these advances, AI has improved significantly in numerous tasks typically performed by humans. These tasks range from competition math to image generation, with some basic metrics exceeding the human baseline by 2020. Artificial intelligence systems are increasingly achieving performance levels comparable to humans, even on complex and analytically demanding tasks. \cite{25}.

However, despite their capabilities, large language models (LLMs) exhibit notable limitations when applied to scientific research ideation and other involved tasks. Most LLMs are trained broadly across a wide array of internet text, making them effective for general understanding that is often too shallow for domain-specific reasoning \cite{KOCON2023101861}. They also struggle to accommodate highly variable, user-specific queries that are common in research contexts. Furthermore, their outputs are not grounded in real-time, trusted external sources, which poses challenges for tasks that require up-to-date and verifiable knowledge \cite{cheng2024dated}.
\section{Related Work}
\subsection{AI For Scientific Discovery}
Numerous idea-generation methods exist in academic literature, including iterative idea improvement through a reinforcement learning-like approach, \cite{35,15,baek2024researchagentiterativeresearchidea}, finding links between ideas using open/closed discovery \cite{10}, and idea generation through reasoning \cite{31,36}, with iterative idea generation being the most common. This tool uses an innovative type of literature-based discovery to generate innovative ideas, mixing two common topics that aren’t traditionally combined in literature.

AI has also been used for scientific discovery in many other fields and ways. AI's data annotation capabilities could be applied to annotating proteins with their functions, its data selection capabilities to select rare occurrences from large amounts of data, and its denoising capabilities to increase the quality of scientific data. AI can also run simulations that can be highly accurate in some cases, like when analyzing force fields \cite{wang2023scientific}.

\raggedbottom
\subsection{The AI Scientist \cite{21}}
A prominent example of a related large language model (LLM) agent designed to accelerate machine learning research is "The AI Scientist." While "The AI Scientist" automates peer review by focusing on 500 papers from the ICLR conference, our approach is more comprehensive, utilizing papers from nine different ML conferences. The AI Scientist is also limited to a few different predetermined templates, with specific topics like grokking and 2D diffusion, while the Budget AI Researcher is able to use all the ideas in 9 top ML conferences as a guide for ideation. This broader scope allows our agent not only to facilitate peer review but also to support idea generation, question answering, and paper categorization, among other tasks. Moreover, while “The AI Scientist” may rely heavily on LLMs for idea generation—potentially compromising accuracy—our agent enhances idea generation by incorporating not only ML conference papers but also their references, leading to more robust outcomes. Furthermore, while "The AI Scientist" is mainly limited to using paid LLMs like GPT-4 via the ChatGPT platform, our agent delivers comparable idea generation capabilities even when using a free model.

\subsection{Scideator \cite{29}}
Another recent example of a research agent is “Scideator,” a RAG and Semantic Scholar-powered tool for recombining multiple facets of papers in literature. “Scideator” uses a set of papers that the user is interested in and allows the user to customize the content of the new paper. It then searches Semantic Scholar for details on relevant papers, which it then uses to aid in idea generation. It also uses Semantic Scholar to generate an aggregated summary of recent papers to gauge novelty. However, since Semantic Scholar only provides the abstract and details about a paper, not the full paper, “Scideatior” may not retrieve the ideas that are not explained abstract, like experimental details and the ensuing discussion. This may also compromise the accuracy of novelty ratings, which are generated in a similar process. Unlike the Budget AI Researcher, which automatically retrieves large subsets of papers from machine learning conference websites to begin the process, “Scideator” relies on user-inputted papers to start the ideation process, limiting its scope to the user's expertise.

\subsection{Performance Evaluations}
ChatGPT is getting more sophisticated over time, so researchers are constantly evaluating its performance in paper writing. One of these evaluations from 2023 showed that on a scale of 1 through 10, the average quality of ChatGPT-generated papers ranged from 5.13 to 7.08 over 4 iterations of self-reflection and rewording of prompts, where 5.5 is the acceptance benchmark. However, some of this deficit in versions 1-3 can be attributed to nonexistent references and citation inaccuracies. Nevertheless, the literature review needed drastic improvements and minor edits to the data analysis, overall framework (which could technically be supplemented with templates), and overall writing style \cite{40}. Several other benchmarks for evaluating LLM performance have been developed, including the Scholarly Novelty Benchmark, which uses pairwise novelty comparisons on arXiv papers, RAG-Novelty, which assesses novelty based on the existing literature, and large-group expert studies on the effectiveness of LLMs in the ideation process \cite{20,32}.

\section{Methods}
\begin{figure}[H]
 \centering
 \includegraphics
   [width=0.99\hsize]
   {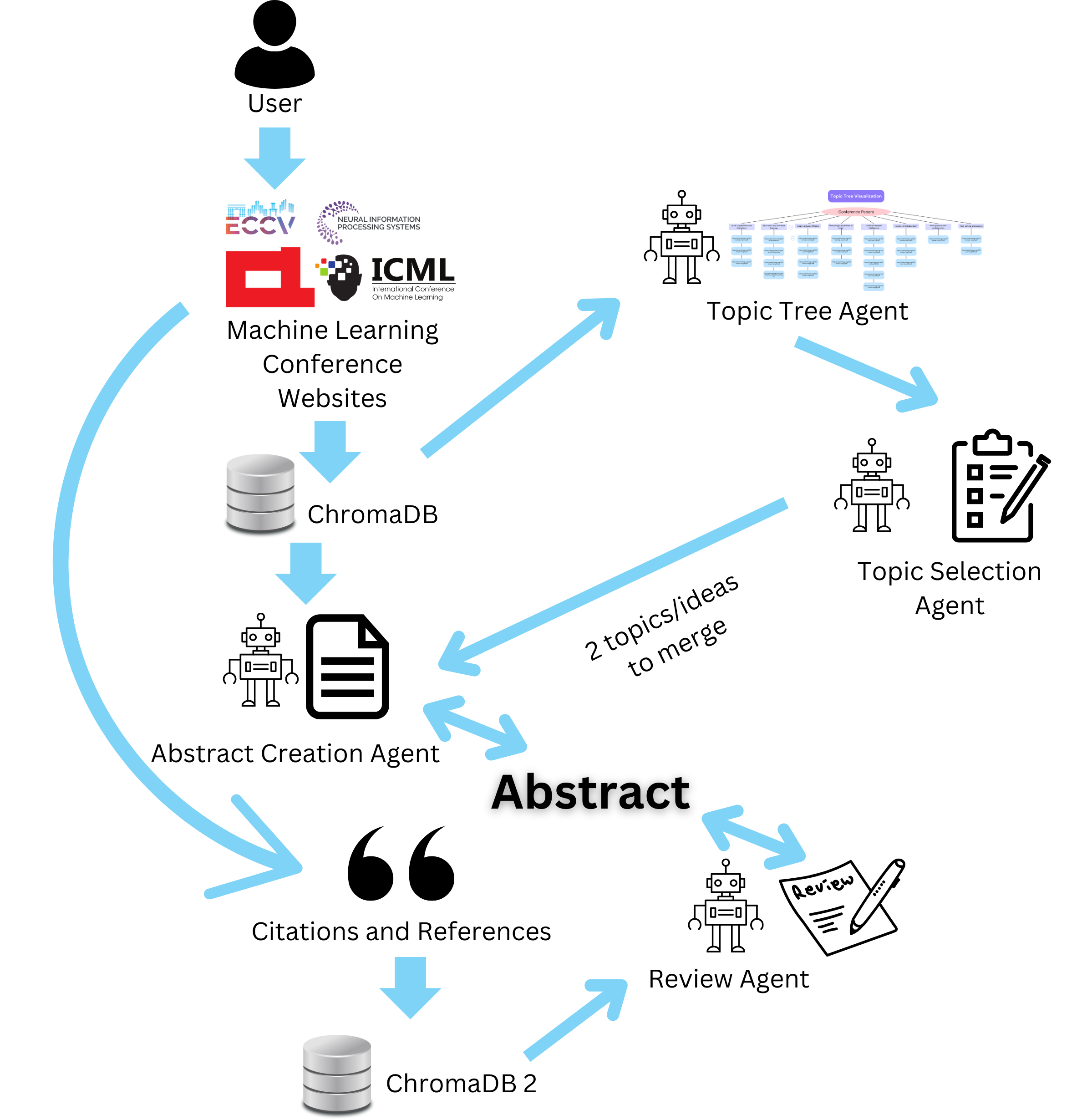}
 \caption{A flowchart describing the overall framework of the Budget AI Researcher }
 \label{fig:framework}
\end{figure}

Traditional large language models (LLMs) often operate as generalists and lack access to up-to-date or domain-specific information. While they can provide general insights, their limitations in their shallow understanding, knowledge cutoffs, and adaptability to users make them insufficient for generating novel and contextually grounded research ideas. To overcome these challenges, our framework fully integrates retrieval-augmented generation (RAG) chains, which incorporates relevant, recent, and personalized information into the generation process, as shown in Fig~\ref{fig:framework}.

RAG chains offer several key advantages that make them well-suited for research ideation. They enable quick and accurate access to specialized knowledge outside the LLM’s training data, such as recent publications, scientific methods, and domain-specific terminology. Their retrieval strategies are customizable, allowing them to align with user intent, topic structures, and input formats. Most importantly, RAG improves factual grounding by linking generated content directly to source documents, an essential feature for producing scientifically credible outputs.

The Budget AI Researcher is designed as a topic-guided ideation agent that generates and refines research abstracts using information retrieved from real-world machine learning papers. This section describes the full system architecture, beginning with the construction of a knowledge base from the nine major AI conferences. We then outline the processing pipeline used to generate a hierarchical topic tree, the pairing of distant topics for ideation, and the refinement of outputs through LLM-based evaluation. In addition, we describe auxiliary features such as question answering and summarization, which enhance user interaction and provide further utility beyond abstract generation.

\subsection{Background}
\subsubsection{LangChain}
LangChain is a tool for applications that chain multiple AI-powered models together. It is mainly used to augment LLMs’ knowledge bases through a RAG (Retrieval Augmented Generation) chain system, multi-agent workflows that tackle a common task, and evaluating results \cite{18}.

\subsubsection{Chroma and Vector Database}

Chroma \cite{39} is an AI-native open-source vector database designed for efficient storage and retrieval of vector embeddings. Vector databases have played critical roles in the era of large language models to support knowledge storage and search and make it easy to build LLM apps. Specifically, Chroma provides tools to embed documents into vectors, to store embeddings as well as their metadata, and to search embeddings. 
To determine the degree of similarity between two vectorized documents, Chroma takes the cosine similarity, Euclidean distance, or inner product between two vector embeddings \cite{17}. 


\subsection{Context Retrieval}
The resources of the ideas for Budget AI Researcher are coming from content retrieved from the nine major AI conferences, the CVPR, ECCV, ICCV, NeurIPS, ICLR, ICML, ACL, EMNLP, and NAACL, using the Python “requests” module. We first extract all paper titles from the conference, and we break them into chunks of 250, which are inputted into Groq \cite{38}. The LLM will then select the 25 most relevant and novel papers from those 250 based on their titles. We then repeat the process for all 250-paper chunks, forming a list of papers to include in the context. We post requests to the accepted paper list of all nine websites, and the resulting source code from each website is retrieved. The source code is then analyzed using BeautifulSoup, which extracts links corresponding to the papers from the website \cite{30}. Depending on the machine learning conference, the agent has to “click” up to three links to get to the PDF, which is done by repeating the requests post to every link BeautifulSoup extracted. If an OpenReview is available, as in the case with NeurIPS and the ICLR, we extract the reviews from those OpenReview sites and store them into a separate Chroma database. Since requests could hit errors when extracting the PDF, we use the “Retry” function, which retries the request up to 5 times if an error is hit. The text is then extracted from the PDFs using PyPDF2 \cite{keawmanee_welcome_2024}.
\subsection{Vectorizing the Text}
The extracted text is split using LangChain’s RecursiveTextSplitter into 3000-character documents. This splitting makes vectorizing large amounts of text easier. The resulting vector embeddings are added to a main collection in a Chroma database \cite{39}, which is then stored in the client’s computer.

\subsection{Topic Tree Generation}
\begin{figure*}[!htbp]
 \centering
 \includegraphics
   [width=0.99\hsize]
   {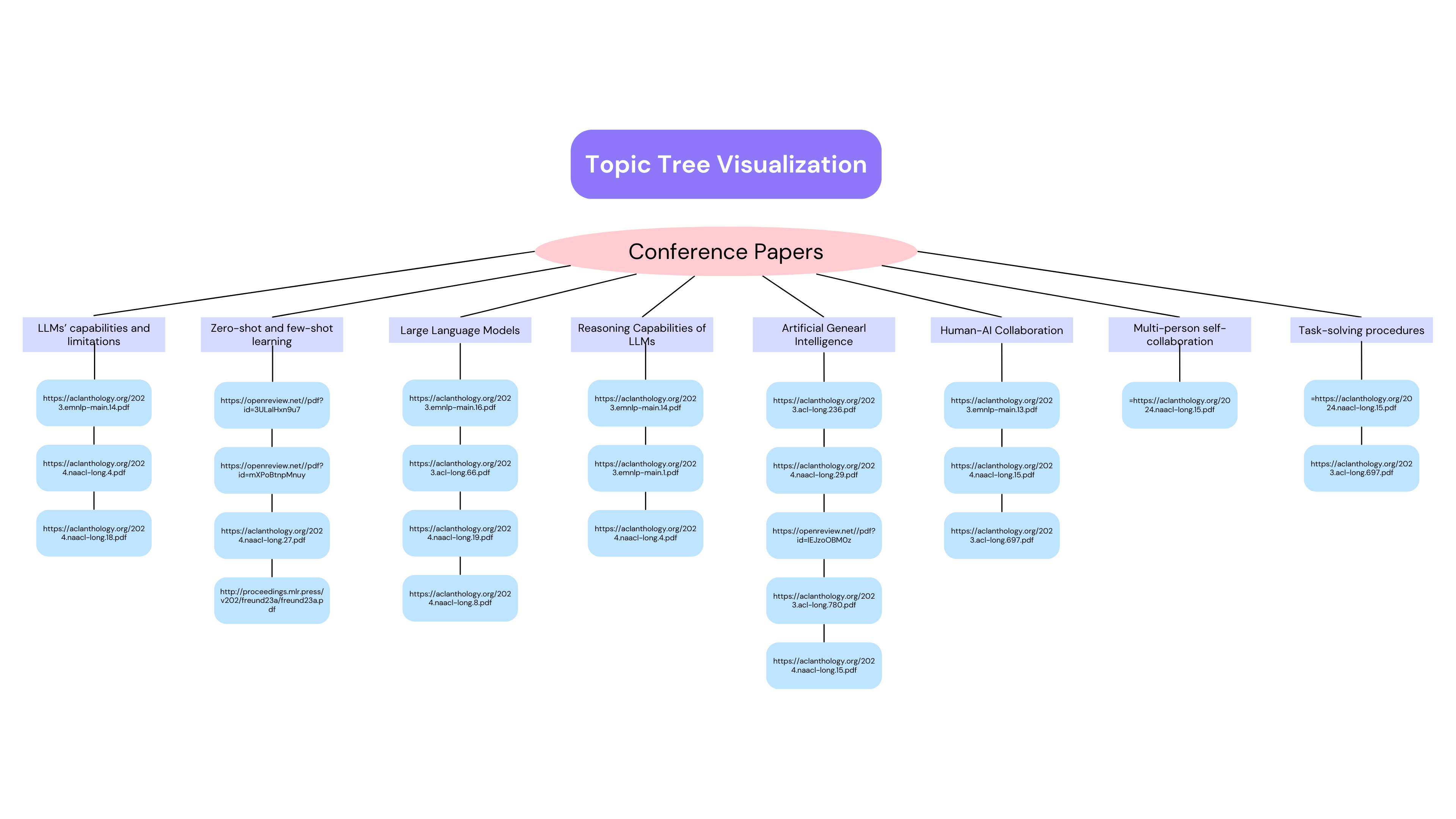}
 \caption{A visualization of a topic tree}
\label{fig:topic-tree}
 \end{figure*}

Topic trees are lists of topics that contain papers as 'children', as shown in Fig~\ref{fig:topic-tree}. They classify papers from machine learning conferences into a few broad categories and provide example papers under each category. Topic trees are useful because they map many key ideas in the AI domain down to a few papers, making it easier to find ways to recombine those ideas for idea generation.

The Budget AI Researcher categorizes the papers by prompting the Llama 3.1 70b-Versatile \cite{22} LLM to generate the 5 most frequent topics using the research papers in the main database. We then use LangChain’s retrieval chain feature to prompt the same LLM to describe each topic \cite{18,24}. In the process, we can retrieve all sources that the LLM uses to generate the description, which comprises all relevant papers to the topic using ChromaDB's retrieval chain feature \cite{39}. It then stores the links to the papers in a 2-D array, with the first dimension being the topic number and the second storing the link to the papers that represent the topic.

\begin{table}[H]
\label{tab:my-table-1}
\begin{tabular}{|p{8cm}|}
\hline
System Prompt\\\hline
\texttt{You are an assistant for question-answering tasks. "
"Use the following pieces of retrieved context to answer "
"the question. If you don't know the answer, try your best."
"Describe each response in a detailed manner, using as many pieces of evidence as possible."
"\{context\}"}
\\\hline
Client Prompt\\\hline
\texttt{"Describe " + i + " in a roughly 5-sentence paragraph."}\\
Here, $i$ represents the topic in the topic tree.
\\\hline
\end{tabular}
\end{table}

\subsection{Question Answering}
One of the functions of the Budget AI Researcher is to answer users’ questions about machine learning using the papers stored in the Chroma database. This feature is implemented using a RAG chain, which accepts a prompt and retrieves similar documents using vector embeddings to answer the prompt \cite{18}. This tool is more useful for researchers who want to gain knowledge in a field they are interested in.
\subsection{Summarization}
Another function of the Budget AI Researcher is summarizing documents retrieved from the machine learning conferences. This tool directly inputs the complete research papers into the large language model using a StuffChain. We then prompt the model to summarize the research paper in around 100 words \cite{18}.

\subsection{Finding the Pair of Topics to Consider}
The hypothesis is many scientific papers contain multiple topics, and many innovative ideas are from innovative combinations of distant topics. Thus the essential step for generating new ideas is to find candidate topic pairs. 
We use Chroma’s “similarity search with score” function between representative documents from each category to find the pair of topics to consider merging to generate the abstract. We trace the document pair with the highest distance, and therefore lowest similarity, back to the topics they represent. Those will be the topics discussed in the research abstract if the user chooses to have the topic chosen for him/her. If not, research ideas will be generated systematically by mixing and matching all pairs of topics, and the user can select which one to use and polish.

\subsection{Abstract Generation}
The main function of the Budget AI Researcher is to generate valid abstracts for novel research papers. This is done in multiple steps.

\subsubsection{Creating the Initial Abstract}
We create the abstract by re-augmenting the model with the AI conference papers, which contain exemplary abstracts that can serve as an example for the LLM. We prompt the model to create an abstract about the two topics chosen in the previous step and to try to create new ideas if possible.

\begin{table}[H]
\label{tab:my-table-2}
\begin{tabular}{|p{8cm}|}
\hline
Prompt\\\hline
\texttt{"Write a detailed research paper abstract about " + topic$\_$list[highest$\_$distance$\_$pair[0]] + " and " + topic$\_$list[highest$\_$distance$\_$pair[1]] + ". Be sure to generate new ideas in the process. Don't just reuse ideas from the context. Only output the abstract."}
\\\hline
\end{tabular}
\end{table}

\subsubsection{Abstract Evaluation and Polishing}
We extract references and citations from relevant papers using the Semantic Scholar API, to which we input paper titles we stored during the paper extraction step. We extracted both references and citations so that the fundamental and new ideas in a topic are all considered. The Semantic Scholar Paper IDs of all references are then extracted. Since the Budget AI Researcher focuses on writing the abstract, retrieving the abstracts of the references using the API is sufficient. We then create a new Chroma database and store the vectorized reference abstracts in it, which we will then use in a RAG chain to evaluate the validity of the abstract generated in the previous step.

The Budget AI Researcher also uses the abstracts retrieved from the Semantic Scholar API to polish its own abstracts, making them more professional and novel by imitating the style of retrieved abstracts. It also combines its own ideas with with ideas from the retrieved abstracts, increasing novelty as well as professionalism.

After this, the Budget AI Researcher names the top 10 most unique abstracts, which we find through prompting those that are the most different from the context. It then combines all pairs of abstracts, amounting up to $\binom{10}{2}=45$ possible research ideas in total for each prompting.

\begin{table}[H]
\label{tab:my-table-3}
\begin{tabular}{|p{8cm}|}
\hline
Prompt\\\hline
\texttt{"Polish the abstract and the idea presented, and make it more valid if it is not already. Make sure to highly emphasize the details that make the research idea different from what's presented in the provided context, and draw a comparison between this method and methods mentioned in the context. If possible, increase the novelty of the abstract by drawing connections between it and details from the context. Make sure to include a title.\textbackslash n\textbackslash n" + abstract}
\\\hline
\end{tabular}
\end{table}
The context in this case is the abstracts retrieved from the Semantic Scholar API.

The Budget AI Researcher also uses the documents collected from OpenReview pages of the conference papers selected from conferences in which OpenReviews are available, such as the NeurIPS and the ICLR, to generate automated peer reviews. These peer reviews are used in conjunction to the Semantic Scholar abstracts to improve the ideas generated by the Budget AI Researcher.
\section{Experiments}
\label{sec:experiments}

\subsection{Qualitative Evaluation and Case Studies}
We will first use examples to show the generation process and use cases of Budget AI Researcher.
\subsubsection{Summarizing Papers }
Model: Llama-3.1-8b-Instant

\begin{table}[H]
\label{tab:my-table-4}
  \resizebox{\linewidth}{!}{
\begin{tabular}{|p{10cm}|}
\hline
\multirow{2}{*}{Sample Summary 2 (Long) \cite{22}}\\
\\\hline
\\
Gametheoretic models of coalition formation have drawn significant interest in recent years due to their ability to capture meaningful properties of multiagent interactions. Hedonic Games (HGs) model coalition formation of strategic agents guided by their individual preferences. A solution in HGs is a partition of agents (or coalition structure) having desirable properties, which typically stand for stability against some kinds of deviations. Core stability is a fundamental and natural requirement, but it is notably very difficult to achieve, even under the usual assumption of full knowledge of agents' preferences.\\
The authors introduce the concept of "-fractional core-stability, a relaxation of the concept of core-stability, where at most an "-fraction of all possible coalitions is allowed to core-block. This notion bears similarities with PAC-stabilizability, a concept that has been previously studied in the context of HGs. The authors investigate this concept on two fundamental classes of HGs: Simple Fractional and Anonymous.\\
For the classes of Simple Fractional and Anonymous HGs, the authors show that it is possible to efficiently compute an "-FC solution under the uniform distribution, with " sub-exponentially small in the number of agents. In particular, in the case of anonymous HGs, an algorithm computing an "-FC solution under the class of $n$-bounded distributions is presented, where the ratio of the probabilities of extracting any two coalitions is bounded by the parameter $n$.\\
The authors also study the existence of "-FC solutions under arbitrary sampling distributions and show that an "-FC may fail to exist for constant values of ". However, for the aforementioned classes of HGs, the authors prove the existence and efficient computation of "-FC solutions for as small as possible values of ".\\
The authors design algorithms that can handle the possible incomplete knowledge of agents' preferences and use the sampling distribution to learn them in a PAC-learning fashion while maintaining the same guarantees on "with high confidence. The $n$-bounded distribution class includes the uniform distribution as a special case and can be considered a suitable extension of it.\\
\\\hline
\end{tabular}
}
\end{table}

\begin{table}[H]
\label{tab:my-table-5}
  \resizebox{\linewidth}{!}{
\begin{tabular}{|p{10cm}|}
\hline
\multirow{2}{*}{Sample Summary 1 (Short) \cite{22}}\\
\\\hline
\\
Researchers have made significant progress in 3D hand reconstruction from color images, with many methods focusing on predicting 3D positions of sparse keypoints. However, these representations are insufficient for precise hand-object interactions. To address this limitation, recent works model the dense hand surface, often using deformable hand mesh models like MANO. These methods can produce realistic hand meshes but are limited by the resolution of parametric meshes. \\
To improve 3D hand-object reconstruction, researchers propose to combine the advantages of parametric mesh models and signed distance functions (SDFs). SDFs can represent arbitrary details, but lack explicit priors, while parametric models provide strong prior knowledge but have limited shape deformations and mesh resolutions. A new pose-normalized SDF framework is proposed to leverage prior knowledge for hand-object reconstruction from monocular RGB images. This method shows significant improvements over the state of the art on challenging benchmarks like ObMan and DexYCB.\\
\\\hline
\end{tabular}
}
\end{table}
As seen in the example summaries, the LLM summarizes the papers comprehensively, accurately, and in a readable manner, although the summaries sometimes must be longer than 100 words to be effective.

\subsubsection{Answering Questions}
Models: Llama3.1-70b-Versatile (Q1)/Google Gemini (Q2)

\begin{table}[H]
\label{tab:my-table-6}
  \resizebox{\linewidth}{!}{
\begin{tabular}{|p{10cm}|}
\hline
\multirow{2}{*}{Question 1: What is a Binary Neural Network? \cite{22}}\\
\\\hline\\
A Binary Neural Network (BNN) is a type of neural network where weights and activation functions are constrained to just one bit, i.e., binary values, typically +1 or -1. This extreme quantization case replaces heavy matrix computations with bit-wise operations, resulting in significant memory compression and speed-up. BNNs are particularly appealing for resource-constrained environments such as edge devices and mobile applications.\\\\\hline
\multirow{2}{*}{Question 2: What is federated learning? \cite{9}}\\
\\\hline\\
Federated learning is a technique that enables multiple devices to train a shared model without sharing private data. Each device trains a local model using its own data, and then the local models are aggregated to create a global model.\\
\\\hline
\end{tabular}
}
\end{table}

Both the Llama 3.1 and Gemini models answer the question concisely and professionally based on the conference papers provided. The answers are also comprehensible to a wide audience.

\subsubsection{Generating Topic Trees}
A sample topic list and topic categorization are shown below.\\\\
\noindent Model: Llama3.3-70b-Versatile
\begin{table}[H]
\label{tab:my-table-7}
  \resizebox{\linewidth}{!}{
\begin{tabular}{|p{10cm}|}
\hline
\multirow{2}{*}{Sample Topic List \cite{22}}\\
\\\hline
\\
Here are the top 10 AI-related topics that appear in the context:

1. Chain-of-thought (CoT) reasoning\\
2. Language model evaluation and benchmarking\\
3. Prompt engineering\\
4. Open-ended problem-solving\\
5. Scientific innovation and discovery\\
6. Large language models (LLMs)\\
7. Cognitive psychology for LLMs\\
8. Adversarial attacks and optimization\\
9. Natural language processing and generation\\
10. Conversational AI and dialogue systems\\
\\\hline
\end{tabular}
}
\end{table}

The AI model-generated topics cover some of the largest topics in AI research. The Llama3.1-70b-Versatile model, paired with Chroma RAG retrieval, will then match research papers to these categories. To do so, we prompted the LLM to generate 5-sentence descriptions of each topic and used the LangChain Question-Answer Chain paired with an RAG Chain to retrieve the sources of information. Due to the limited context window as of now, as Groq limited context windows to 8000 tokens for models capable of retrieving from contexts of up to 128000 tokens like the Llama 3.1-3.2 models \cite{38}, the papers extracted may be concentrated toward a few conferences.

The sample topic tree is shown in Appendix E.

\subsubsection{Generating Abstracts }

\begin{table}[H]
\label{tab:my-table-8}
  \resizebox{\linewidth}{!}{
\begin{tabular}{|p{10cm}|}
\hline
\multirow{2}{*}{Sample Abstract \cite{22}}\\
\\\hline
\\
Evaluating the Effectiveness of ChatGPT in Assessing Answer Correctness: A Study on Large Language Models
\\
Abstract:\\
This paper presents a comprehensive evaluation of ChatGPT's ability to assess the correctness of answers generated by large language models (LLMs). We propose a novel evaluation methodology that utilizes ChatGPT to check the correctness of answers and introduces a new metric, accuracy (ALM), to measure the percentage of correct answers judged by ChatGPT. Our results show a high agreement (98$\%$) between human judgment and LLM-based metrics, demonstrating the reliability of ChatGPT in evaluating answer correctness. We also investigate the impact of prompting LLMs to respond "unsure" when not confident in the answer, which significantly reduces hallucination rates. Our findings have implications for the development of more accurate and reliable LLMs and highlight the potential of ChatGPT as a tool for evaluating answer correctness.\\
\\\hline
\end{tabular}
}
\end{table}

The above abstract merges the ideas of LLM accuracy, and evaluation methodology, which could help researchers find new connections between ideas. These topics were chosen using the maximum distance score between the abstract of a paper in one topic and that of a paper in another topic across all pairs of topics from the topic tree. Since this method may result in the same pairs of topics being picked across multiple trials, the Budget AI Researcher may also select two topics from the topic tree at random and merge them together to create a new idea. Since these merges may result in vague or infeasible ideas, the Budget AI Researcher can generate an experimental procedure, a sample of which is in Appendix F.

\subsubsection{Evaluating Abstracts Using Semantic Scholar }
Model: Llama-3.1-70b-Versatile

\begin{table}[H]
\label{tab:my-table-9}
  \resizebox{\linewidth}{!}{
\begin{tabular}{|p{10cm}|}
\hline
\multirow{2}{*}{Sample Abstract Evaluation \cite{22}}\\
 \\
 \hline\\
The provided abstract is valid based on the given papers. The abstract discusses the evaluation of ChatGPT's ability to assess answer correctness, reduction of hallucinations, and improvement of large language models (LLMs) through external feedback and tools. These topics are consistent with the themes and ideas presented in the provided papers, such as the Holistic Evaluation of Language Models (HELM), LLM-Augmenter system, and CRITIC framework.\\
\\\hline
\multirow{2}{*}{Sample Polished Abstract \cite{22}}\\
\\\hline\\
This paper presents a comprehensive evaluation of ChatGPT's ability to assess the correctness of answers generated by large language models (LLMs), introducing a novel evaluation methodology and the accuracy (ALM) metric. Our results demonstrate a high agreement between human judgment and LLM-based metrics, showcasing ChatGPT's reliability. By leveraging external knowledge and iterative revisions, this research explores innovative methods to improve LLM performance, such as the LLM-Augmenter system and the CRITIC framework, which validate and refine outputs using external tools, ultimately enhancing the accuracy and reliability of LLMs.\\
\\\hline
\end{tabular}
}
\end{table}

The polished abstract uses more examples, such as the CRITIC framework and the LLM-Augmenter system, to back up its validity, and it appears more professional than the initial abstract.

\subsubsection{Evaluating Abstracts using OpenReview}
The Budget AI Researcher uses the OpenReview pages of selected papers from conferences such as the ICLR and NeurIPS stored in the Reviewer Chroma database to evaluate the ideas it generated and improve its ideas. A sample review is shown below.
\begin{table}[H]
    \centering
    \resizebox{\linewidth}{!}{
    \begin{tabular}{|p{10cm}|}
    \hline\\
    Sample Review based on Collected OpenReview Data\\\\
    \hline\\
         **Review Summary:**

        **Summary:**  
        The paper addresses the critical intersection of ethics, safety, and human cognition in AI development. It introduces a framework to mitigate biases and promote ethical AI behavior through transparency and accountability.
        
        **Strengths:**  
        The research is well-motivated, addressing a crucial area in AI ethics. The proposed framework is comprehensive and offers a clear path for ensuring alignment with human values.
        
        **Weaknesses:**  
        The paper may lack empirical validation and real-world application examples. It could benefit from comparing its approach with existing methods and including user studies to assess the framework's practical impact.
        
        **Conclusion:**  
        While the framework is promising, further empirical support and comparative analysis would strengthen its contribution.
        \\\\\hline
    \end{tabular}
    }
\end{table}

\subsection{Quantitative Evaluation}
\begin{table}[H]
\caption{	Performance of in Abstract-Generation Tasks (October 4, 2024-March 16, 2025).
}
\label{tab:my-table-10}
  \resizebox{\linewidth}{!}{
\begin{tabular}{ccccc}
\hline\\
\multirow{2}{*}{Category}	 & \multicolumn{3}{c}{Large Language Model}&\\\\
	&GPT-4o-mini	&Llama 3.2 90B	&Claude	&The Budget AI Researcher\\
 \\\hline
\\
Interestingness&	8.40	&8.15&	8.05&	8.37\\
Novelty&	7.55&	7.30&	7.30&	8.13\\
Feasibility&	7.80&	7.70	&7.55&	7.89\\
\\\hline
\end{tabular}
}
\label{tb:abstract-gen}
\end{table}
For quantitative evaluation, we evaluate the generated ideas from three perspectives (interestingness, novelty, and feasibility) using one or few-shot learning through ChatGPT-4o using examples and ratings from reference \cite{21}. Table~\ref{tb:abstract-gen} shows the average rating of sample abstracts generated from each framework. The baseline LLMs, which have knowledge cutoffs of October 2023, December 2023, and April 2024 for GPT-4o-mini \cite{BibEntry2024Nov2}, Llama 3.2 90B \cite{BibEntry2024Nov3}, and Claude Sonnet 3.5 \cite{BibEntry2024Nov} respectively, were each asked to generate research paper abstracts that were interesting and novel, and the Budget AI Researcher generated and polished abstracts for distant combinations of topics from the topic tree. The Budget AI Researcher, which uses Llama 3.2 11B Vision, generates comparably interesting, much more novel, and comparably feasible ideas than those generated by leading LLMs. We performed an updated round of ChatGPT evaluation on March 16, 2025.

\begin{table}[H]
\caption{		Probability of Generating Similar Ideas to Those of Various 2024 ML Conferences.
}
\label{tab:my-table-11}
  \resizebox{\linewidth}{!}{
\begin{tabular}{ccc cc}
\hline\\
\multirow{2}{*}{Category}	 & \multicolumn{3}{c}{Large Language Model}&\\\\
	&GPT-4o-mini	&Llama 3.2 90B	&Claude	&The Budget AI searcher   \\    \\\hline\\
ECCV, NeurIPS, ICML, ICLR, ACL &0.52&0.35&0.50&0.60\\
\\\hline
\end{tabular}
}
\label{tb:prob}
\end{table}

We further evaluate the generated ideas using "future publications" that an LLM does not have knowledge of. Specifically, we use the Llama 3.1 8B Instant model augmented with 200 papers from the 2024 ECCV, NeurIPS, ICML, ICLR, and ACL, amounting to 1000 papers total. When the Budget AI Researcher was first developed, these five conferences hadn’t convened in 2024 yet, which makes comparing ideas from different agents against those presented in the 2024 cycle of these conferences an objective performance metric. This is because the conferences present novel ideas as of the time they were held.

The augmented Llama 3.1 8B model rates the similarity from 0 to 1 between the 1000 papers in context and the 20 sample abstracts generated by each LLM/agent individually. The average ratings for samples from each LLM/agent are recorded in Table~\ref{tb:prob}. The Budget AI Researcher, using data from the 2023 iteration of these five conferences (and four extra conferences whose 2024 iteration had already occurred before the development of the Budget AI Researcher) comes up with much more similar ideas to the 2024 iterations of the same conferences than leading LLMs do. These data further underscore the novelty and feasibility of the ideas that the Budget AI Researcher generates. 

\begin{table}[H]
\caption{	Performance of in Abstract-Generation Tasks.
}
\label{tab:my-table-12}
  \resizebox{\linewidth}{!}{
\begin{tabular}{ccc}
\hline \\
Category	 &\multicolumn{2}{c}{Large Language Model}\\\\
	&Augmented Llama 3.2 11B&	The Budget AI Researcher 
           \\  \\\hline\\
Interestingness	&7.30&	8.37\\
Novelty&	6.55	&8.13\\
Feasibility	&7.90	&7.89  \\
\\\hline
\end{tabular}
}
\end{table}

\subsection{Human Evaluation} In addition to using ChatGPT with one-shot learning as a judge, we performed a human evaluation on sample abstracts from the Budget AI Researcher and the AI Scientist, which were randomly mixed, deidentified, and given to 6 evaluators, who rated the ideas from 1-5. The evaluators are current master's and Ph.D. students at SUNY Stony Brook University who are not related to this project. The abstracts from the Budget AI Researcher were generated after one iteration of ideation (mixing and matching ideas). We generated the abstracts from the AI Scientist by using the ideation template from 2D Diffusion and the OpenAI API. We averaged the results for the ideas from the Budget AI Researcher and the AI Scientist separately for each evaulator, and we averaged those results to get our final figures, shown in the table below.

\begin{table}[H]
\caption{Human Evaluations Compared to the AI Scientist}
\label{tab:my-table-13}
  \resizebox{\linewidth}{!}{
\begin{tabular}{ccc}
\hline \\
Category	 &\multicolumn{2}{c}{Framework}\\\\
	&The AI Scientist & The Budget AI Researcher 
           \\  \\\hline\\
Interestingness	&2.925&	3.583\\
Feasibility&	3.575	&3.550\\
Novelty	&3.275	&3.233  \\
\\\hline
\end{tabular}
}
\end{table}

Evidently, the interestingness of ideas generated by the Budget AI Researcher is better than that of ideas generated by the AI Scientist, and the feasibility and novelty are comparable.

\subsection{Ablation Study} To demonstrate the usefulness of the topic guide module, we compare the Budget AI Researcher with its backbone model, a plain augmented Llama 3.2 11B LLM, removing the topics in the prompt. Budget AI Researcher largely outperforms the ablation model in interestingness and novelty. The lower feasibility of the ideas generated by the Budget AI Researcher comes from the fact that recombining ideas that are far apart can lead to research goals that are harder to achieve.
\section{Discussion }
\subsection{2024 ICML Evaluation}
Using a RAG-augmented LLM, we can evaluate the similarity between the abstract presented in the previous section and the ideas presented in 2024 ICML papers. The sample and similarity evaluation are in Appendix A.

The abstract generated by the Budget AI Researcher in the previous section, trained on ML conferences held in 2023, shared common themes as some of the papers accepted into the 2024 ICML conference, highlighting its ability to foresee advancements.
\subsection{ChatGPT Evaluation $\&$ Improvements}
We used ChatGPT 4o to evaluate the originality and applicability of research ideas presented by the Budget AI Researcher. The evaluation of the previous idea is shown in Appendix B.

We enhanced the abstract review and refinement process by instructing the LLM to elaborate on the unique aspects of the abstract, which are not present in the given context. The improvement in ChatGPT evaluation is shown in Appendix C.

We can further polish the abstract using 2024 ML conference papers. In this experiment, we augmented the Llama 3.2-11b-Vision model with 400 papers from the ICML 2024 and prompted it to freely add details to make the abstract more novel. A repolished abstract using these features is included in Appendix D. 

Considering the feedback from ChatGPT, the Budget AI Researcher does quite well in generating novel and practical ideas, even though the methods for doing so are simpler than those employed in similar applications. By proposing unique applications of concepts like ALM to aid in LLM self-evaluation and hallucination prevention, the Budget AI Researcher generates publishable topics and ideas for researchers to explore.
\subsection{Limitations \& Future Work}
Although the Budget AI Researcher is a comprehensive research tool, it still has limitations. It is bound to a set of thousands of papers from 9 conferences and many of their citations, and while this encompasses most of the subjects in AI, it is far from what the entire Internet has to offer.

Since rate limits on the Groq API (typically 6000 tokens) can only encompass small bits and pieces of the conference papers, the LLM may not be able to retrieve all the experimental procedures required. This could lead to the experimental procedures to achieve the goals outlined in the abstracts being vague \cite{38}.

To resolve these limitations, we can add an online search feature such as the Semantic Scholar API relevance search to strengthen the ideation process. To generate finer-grained ideas, we can employ chain-of-thought with a modified topic tree, where each paper in the tree would be broken down into the task, the method, and the peer review if it is available. To resolve the limitation of rate limits, we can run the LLMs locally instead of using the Groq API.

\section{Conclusion}
The Budget AI Researcher is a free tool designed to reduce the time researchers spend reading papers and brainstorming practical and interesting ideas. It leverages LLMs and retrieval-augmented generation (RAG) with knowledge drawn from the nine major machine learning conferences to support tasks such as summarization, abstract generation and refinement, and paper categorization. By grounding its outputs in real-world research papers and recombining ideas using a topic-guided structure, the system significantly accelerates the early stages of the research process in a cost-free, accessible way.

This work also inspires solutions to a broader and increasingly critical challenge across domains: how to generate personalized, context-aware outputs grounded in diverse, evolving, and real-world knowledge. The strengths of RAG chains---dynamic context injection, low-latency access to specialized information, customizable retrieval strategies, and improved factual grounding---position this approach as a critical step towards solving this problem.

While this study focuses on research ideation, the underlying methodology has broader potential. The Budget AI Researcher's ability to generate personalized context-sensitive outputs grounded in evolving knowledge makes its structure well suited for domains such as experimental design, personalized learning, and dynamic role-playing in therapeutic contexts. These applications share a common need for contextual reasoning informed by external, updated-in-real-time information, making RAG-powered systems a promising foundation for future development.

\bibliography{aaai25.bib}

\begin{thebibliography}{31}
\providecommand{\natexlab}[1]{#1}

\bibitem[{Baek et~al.(2024)Baek, Jauhar, Cucerzan, and
  Hwang}]{baek2024researchagentiterativeresearchidea}
Baek, J.; Jauhar, S.~K.; Cucerzan, S.; and Hwang, S.~J. 2024.
\newblock ResearchAgent: Iterative Research Idea Generation over Scientific
  Literature with Large Language Models.
\newblock Available: \url{https://arxiv.org/abs/2404.07738}, arXiv:2404.07738.

\bibitem[{Cheng et~al.(2024)Cheng, Marone, Weller, Lawrie, Khashabi, and
  Van~Durme}]{cheng2024dated}
Cheng, J.; Marone, M.; Weller, O.; Lawrie, D.; Khashabi, D.; and Van~Durme, B.
  2024.
\newblock Dated Data: Tracing Knowledge Cutoffs in Large Language Models.
\newblock \emph{arXiv preprint arXiv:2403.12958}.
\newblock Version 2, revised on 17 Sep 2024. Available at:
  \url{https://arxiv.org/abs/2403.12958}.

\bibitem[{Dubey et~al.(2024)Dubey, Jauhri, Pandey, Kadian, Al-Dahle, Letman,
  Mathur, Schelten, Yang, Fan et~al.}]{22}
Dubey, A.; Jauhri, A.; Pandey, A.; Kadian, A.; Al-Dahle, A.; Letman, A.;
  Mathur, A.; Schelten, A.; Yang, A.; Fan, A.; et~al. 2024.
\newblock The llama 3 herd of models.
\newblock \emph{arXiv preprint arXiv:2407.21783}.

\bibitem[{Dwivedi et~al.(2019)Dwivedi, Hughes, Ismagilova, Aarts, Coombs,
  Crick, Duan et~al.}]{8}
Dwivedi, Y.~K.; Hughes, L.; Ismagilova, E.; Aarts, G.; Coombs, C.; Crick, T.;
  Duan, Y.; et~al. 2019.
\newblock Artificial Intelligence (AI): Multidisciplinary Perspectives on
  Emerging Challenges, Opportunities, and Agenda for Research, Practice and
  Policy.
\newblock \emph{International Journal of Information Management}, 57: 101994.

\bibitem[{et~al.(2023)}]{KOCON2023101861}
et~al., J.~K. 2023.
\newblock ChatGPT: Jack of all trades, master of none.
\newblock \emph{Information Fusion}, 99: 101861.

\bibitem[{{Groq, Inc.}(2024)}]{38}
{Groq, Inc.} 2024.
\newblock GroqCloud.
\newblock Available: \url{https://console.groq.com/docs/models}.

\bibitem[{Henry and McInnes(2017)}]{10}
Henry, S.; and McInnes, B.~T. 2017.
\newblock Literature Based Discovery: Models, Methods, and Trends.
\newblock \emph{Journal of Biomedical Informatics}, 74: 20--32.

\bibitem[{Hubbard and Dunbar(2017)}]{12}
Hubbard, K.~E.; and Dunbar, S.~D. 2017.
\newblock Perceptions of Scientific Research Literature and Strategies for
  Reading Papers Depend on Academic Career Stage.
\newblock \emph{PLOS ONE}, 12(12): e0189753.

\bibitem[{Ifargan et~al.(2024)Ifargan, Hafner, Kern, Alcalay, and Kishony}]{15}
Ifargan, T.; Hafner, L.; Kern, M.; Alcalay, O.; and Kishony, R. 2024.
\newblock Autonomous LLM-driven research from data to human-verifiable research
  papers.
\newblock Available: \url{https://arxiv.org/abs/2404.17605}, arXiv:2404.17605.

\bibitem[{Jiang et~al.(2024)Jiang, Sablayrolles, Roux, Mensch, Savary, Bamford,
  Chaplot, Casas, Hanna, Bressand et~al.}]{24}
Jiang, A.~Q.; Sablayrolles, A.; Roux, A.; Mensch, A.; Savary, B.; Bamford, C.;
  Chaplot, D.~S.; Casas, D. d.~l.; Hanna, E.~B.; Bressand, F.; et~al. 2024.
\newblock Mixtral of experts.
\newblock \emph{arXiv preprint arXiv:2401.04088}.

\bibitem[{Keawmanee(2024)}]{keawmanee_welcome_2024}
Keawmanee, N. 2024.
\newblock Welcome to {PyPDF2}
  ıfmmode---{\textbackslash}else---{\textbackslash}fi {PyPDF2} documentation.
\newblock Available: \url{https://pypdf2.readthedocs.io/en/3.x}.

\bibitem[{Kedia(2024)}]{39}
Kedia, S. 2024.
\newblock Chroma.
\newblock Available: \url{https://www.trychroma.com}.

\bibitem[{Khlaif et~al.(2023)Khlaif, Mousa, Hattab, Itmazi, Hassan, Sanmugam,
  and Ayyoub}]{40}
Khlaif, Z.~N.; Mousa, A.; Hattab, M.~K.; Itmazi, J.; Hassan, A.~A.; Sanmugam,
  M.; and Ayyoub, A. 2023.
\newblock The Potential and Concerns of Using AI in Scientific Research:
  ChatGPT Performance Evaluation.
\newblock \emph{JMIR Medical Education}, 9: e47049.

\bibitem[{Kukreja et~al.(2023)Kukreja, Kumar, Bharate, Purohit, Dasgupta, and
  Guha}]{17}
Kukreja, S.; Kumar, T.; Bharate, V.; Purohit, A.; Dasgupta, A.; and Guha, D.
  2023.
\newblock Vector Databases and Vector Embeddings-Review.
\newblock In \emph{2023 International Workshop on Artificial Intelligence and
  Image Processing (IWAIIP)}, 231--36. IEEE.

\bibitem[{Lin, Peng, and Fang(2024)}]{20}
Lin, E.; Peng, Z.; and Fang, Y. 2024.
\newblock Evaluating and Enhancing Large Language Models for Novelty Assessment
  in Scholarly Publications.
\newblock Available: \url{https://arxiv.org/abs/2409.16605}, arXiv:2409.16605.

\bibitem[{Lu et~al.(2024)Lu, Lu, Lange, Foerster, Clune, and Ha}]{21}
Lu, C.; Lu, C.; Lange, R.~T.; Foerster, J.; Clune, J.; and Ha, D. 2024.
\newblock The AI Scientist: Towards Fully Automated Open-Ended Scientific
  Discovery.
\newblock Available: \url{https://arxiv.org/abs/2408.06292}, arXiv:2408.06292.

\bibitem[{Nassej(2024)}]{25}
Nassej, M. 2024.
\newblock Artificial Intelligence Index Report 2024.
\newblock Technical report.
\newblock Accessed: Feb. 22, 2025. [Online]. Available:
  \url{https://aiindex.stanford.edu/wp-content/uploads/2024/04/HAI_2024_AI-Index-Report.pdf}.

\bibitem[{OpenAI(2024a)}]{27}
OpenAI. 2024a.
\newblock ChatGPT.

\bibitem[{{Paperpile}(2024)}]{11}
{Paperpile}. 2024.
\newblock {How to Read a Scientific Paper [3 Steps - 2024]}.
\newblock Accessed: Feb. 22, 2025. [Online]. Available:
  \url{https://paperpile.com/g/read-scientific-paper}.

\bibitem[{PromptHub(2024{\natexlab{a}})}]{BibEntry2024Nov}
PromptHub. 2024{\natexlab{a}}.
\newblock {Claude 3.5 Sonnet Model Card}.
\newblock [Online; accessed 24. Nov. 2024]. Available:
  \url{https://www.prompthub.us/models/claude-3-5-sonnet}.

\bibitem[{PromptHub(2024{\natexlab{b}})}]{BibEntry2024Nov2}
PromptHub. 2024{\natexlab{b}}.
\newblock {GPT-4o mini Model Card}.
\newblock [Online; accessed 24. Nov. 2024]. Available:
  \url{https://www.prompthub.us/models/gpt-4o-mini}.

\bibitem[{PromptHub(2024{\natexlab{c}})}]{BibEntry2024Nov3}
PromptHub. 2024{\natexlab{c}}.
\newblock {Llama 3.2 90B Model Card}.
\newblock [Online; accessed 24. Nov. 2024]. Available:
  \url{https://www.prompthub.us/models/llama-3-2-90b}.

\bibitem[{Radensky et~al.(2024)Radensky, Shahid, Fok, Siangliulue, Hope, and
  Weld}]{29}
Radensky, M.; Shahid, S.; Fok, R.; Siangliulue, P.; Hope, T.; and Weld, D.~S.
  2024.
\newblock Scideator: Human-LLM Scientific Idea Generation Grounded in
  Research-Paper Facet Recombination.
\newblock Available: \url{https://arxiv.org/abs/2409.14634}, arXiv:2409.14634.

\bibitem[{Richardson(2021)}]{30}
Richardson, L. 2021.
\newblock Beautiful Soup Documentation Ifmmode---Beautiful Soup 4.4.0
  Documentation.
\newblock Available: \url{https://beautiful-soup-4.readthedocs.io/en/latest}.

\bibitem[{Sel et~al.(2024)Sel, Al-Tawaha, Khattar, Jia, and Jin}]{31}
Sel, B.; Al-Tawaha, A.; Khattar, V.; Jia, R.; and Jin, M. 2024.
\newblock Algorithm of Thoughts: Enhancing Exploration of Ideas in Large
  Language Models.
\newblock Available: \url{https://arxiv.org/abs/2308.10379}, arXiv:2308.10379.

\bibitem[{Si, Yang, and Hashimoto(n.d.)}]{32}
Si, C.; Yang, D.; and Hashimoto, T. n.d.
\newblock Can LLMs Generate Novel Research Ideas?

\bibitem[{Team et~al.(2023)Team, Anil, Borgeaud, Alayrac, Yu, Soricut,
  Schalkwyk, Dai, Hauth, Millican et~al.}]{9}
Team, G.; Anil, R.; Borgeaud, S.; Alayrac, J.-B.; Yu, J.; Soricut, R.;
  Schalkwyk, J.; Dai, A.~M.; Hauth, A.; Millican, K.; et~al. 2023.
\newblock Gemini: a family of highly capable multimodal models.
\newblock \emph{arXiv preprint arXiv:2312.11805}.

\bibitem[{Topsakal and Akinci(2023)}]{18}
Topsakal, O.; and Akinci, T.~C. 2023.
\newblock Creating large language model applications utilizing langchain: A
  primer on developing llm apps fast.
\newblock In \emph{International Conference on Applied Engineering and Natural
  Sciences}, volume~1, 1050--1056.

\bibitem[{Wang et~al.(2023)Wang, Fu, Du, Gao, Huang, Liu, Chandak, Liu,
  Van~Katwyk, Deac et~al.}]{wang2023scientific}
Wang, H.; Fu, T.; Du, Y.; Gao, W.; Huang, K.; Liu, Z.; Chandak, P.; Liu, S.;
  Van~Katwyk, P.; Deac, A.; et~al. 2023.
\newblock Scientific discovery in the age of artificial intelligence.
\newblock \emph{Nature}, 620(7972): 47--60.

\bibitem[{Wang et~al.(2024)Wang, Downey, Ji, and Hope}]{35}
Wang, Q.; Downey, D.; Ji, H.; and Hope, T. 2024.
\newblock SciMON: Scientific Inspiration Machines Optimized for Novelty.
\newblock Available: \url{https://arxiv.org/abs/2305.14259}, arXiv:2305.14259.

\bibitem[{Yao et~al.(2023)Yao, Yu, Zhao, Shafran, Griffiths, Cao, and
  Narasimhan}]{36}
Yao, S.; Yu, D.; Zhao, J.; Shafran, I.; Griffiths, T.~L.; Cao, Y.; and
  Narasimhan, K. 2023.
\newblock Tree of Thoughts: Deliberate Problem Solving with Large Language
  Models.
\newblock Available: \url{https://arxiv.org/abs/2305.10601}, arXiv:2305.10601.

\end{thebibliography}

\newpage

\appendices
\section{Appendix A}
Model: Llama3.1-70b-Versatile
\begin{table}[H]
\label{tab:my-table-14}
  \resizebox{\linewidth}{!}{
\begin{tabular}{|p{10cm}|}
\hline
\multirow{2}{*}{Similar Papers Published in 2024 (“ICML 2024 Papers” 2024)}\\
\\\hline
\\
The papers that are most relevant to the abstract are:
\\
1. "Hallucination Snowballing: An Investigation of Over-commitment in Language Model Hallucinations" - This paper shares the context's focus on language model hallucinations and investigates a similar phenomenon.
\\
2. "Linguistic Calibration for Long-form Generations in Language Models" - This paper discusses the importance of linguistic calibration in language models, which is related to the abstract's mention of conveying the probability of claims.
\\
3. "GRAdual self-truThifying (GRATH): A Post-processing Method for Enhancing Truthfulness in Language Models" - This paper proposes a method for improving truthfulness in language models, which is directly relevant to the abstract's focus on truthfulness.
\\
4. "Evaluating Counterfactual Simulatability of Natural Language Explanations from Language Models" - This paper introduces the concept of counterfactual simulatability, which is related to the abstract's discussion of how language models can help humans build mental models.
\\
5. "Understanding and Mitigating Hallucinations in Large Language Models" - This paper investigates the underlying mechanisms of hallucinations in language models, similar to the abstract, and proposes a solution for hallucination mitigation.
\\
6. "Permutation Sensitivity in Multiple-choice Question Answering: A Vulnerability in Language and Vision-language Models" - While not directly related to the abstract, this paper discusses a specific vulnerability in popular language models, which is crucial to understand when deploying these models in practice.\\
\\\hline
\end{tabular}
}
\end{table}

\section{Appendix B}
ChatGPT Evaluation \cite{27}
\begin{table}[H]
\label{tab:my-table-15}
  \resizebox{\linewidth}{!}{
\begin{tabular}{|p{10cm}|}
\hline
\multirow{2}{*}{Originality: 6/10}\\
\\\hline\\
•	Strengths: The paper introduces a new evaluation methodology using ChatGPT to assess the correctness of LLM-generated answers and proposes a new metric (accuracy ALM). This approach seems somewhat novel in terms of utilizing an LLM like ChatGPT for evaluation purposes.\\
•	Weaknesses: While the methodology is innovative, the concept of using LLMs for self-evaluation and introducing new metrics is not entirely new. The abstract lacks details on how the proposed metric differs significantly from existing metrics, which slightly diminishes the perceived originality.\\
Practicality: 8/10\\
•	Strengths: The paper's findings are highly practical, particularly the high agreement rate between human judgment and LLM-based metrics. The investigation into prompting LLMs to respond "unsure" when not confident is also very relevant to real-world applications, as it addresses the issue of hallucinations in AI responses.\\
•	Weaknesses: The practicality is strong, but the abstract could benefit from more specific examples or potential applications to emphasize how these findings could be implemented in practice. The abstract's claim of high agreement is promising, but without details on the methodology's application or limitations, it’s harder to fully gauge its real-world impact.\\
Overall, the abstract presents a solid contribution with practical implications, though its originality could be strengthened with more distinct innovations or clearer differentiation from existing approaches.\\
\\\hline
\end{tabular}
}
\end{table}

\section{Appendix C}
\begin{table}[H]
\label{tab:my-table-16}
  \resizebox{\linewidth}{!}{
\begin{tabular}{|p{10cm}|}
\hline
\multirow{2}{*}{Sample Polished Abstract}\\
\\\hline\\
This paper introduces a novel evaluation methodology that leverages ChatGPT to assess the correctness of answers generated by large language models (LLMs). Unlike existing approaches that rely on external knowledge or utility functions, our method utilizes ChatGPT as a self-contained evaluation tool, introducing a new metric, accuracy (ALM), to measure the percentage of correct answers judged by ChatGPT. Our results show a high agreement (98$\%$) between human judgment and LLM-based metrics, demonstrating the reliability of ChatGPT in evaluating answer correctness. In contrast to LLM-Augmenter, which augments LLMs with external knowledge, and HELM, which evaluates LMs on multiple scenarios and metrics, our approach focuses on the inherent ability of ChatGPT to evaluate answer correctness, highlighting its potential as a standalone evaluation tool.
ChatGPT Evaluation (OpenAI 2024a)\\
\\\hline
\multirow{2}{*}{Originality: 7/10}\\
\\\hline\\
•	Strengths: The abstract introduces a novel approach by leveraging ChatGPT itself to evaluate the correctness of answers generated by other large language models (LLMs). This self-referential use of an LLM for evaluation purposes is relatively innovative. Additionally, the development of a new metric, accuracy (ALM), tailored specifically for this evaluation adds to the originality of the work.\\
•	Weaknesses: While the approach is fresh, the concept of using LLMs for evaluation is becoming more common in the field. The abstract could further emphasize what makes the ALM metric distinct from existing metrics or how the methodology uniquely differentiates from other LLM-based evaluation strategies to enhance its originality score.
\\
Practicality: 8/10\\
•	Strengths: The high agreement rate of 98$\%$ between human judgment and the LLM-based metric demonstrates strong reliability, making the methodology highly practical for real-world applications. Using ChatGPT as an evaluation tool can streamline the validation process, reducing the need for extensive human evaluation and accelerating development cycles for LLMs.\\
•	Weaknesses: The abstract could provide more details on specific use cases or applications where this evaluation methodology can be implemented effectively. Additionally, discussing any potential limitations or scenarios where the methodology might face challenges would give a more comprehensive view of its practicality.
\\
Overall, the abstract presents a solid and innovative approach with significant practical implications. Enhancing the discussion on what sets the methodology apart and providing more context on its applications could further strengthen its originality and practicality.\\
\\\hline
\end{tabular}
}
Note: The papers that inspired the abstract were taken from machine learning conferences held in 2023. Newer papers on this subject may have been published already, leading to a loss of originality.
\end{table}

\section{Appendix D}
\raggedbottom
\begin{table}[H]
\label{tab:my-table-17}
  \resizebox{\linewidth}{!}{
\begin{tabular}{|p{10cm}|}
\hline
\multirow{2}{*}{Sample Polished Abstract (Novelty Rating: 8/10)}\\
\\\hline\\
This paper presents a comprehensive evaluation of ChatGPT's ability to assess the correctness of answers generated by large language models (LLMs), tackling a major risk in practical applications: hallucination. We propose a novel evaluation methodology that utilizes ChatGPT to check the correctness of answers and introduces a new metric, accuracy (ALM), to measure the percentage of correct answers judged by ChatGPT. Our results show a high agreement (98$\%$) between human judgment and LLM-based metrics, demonstrating the reliability of ChatGPT in evaluating answer correctness. Additionally, we investigate the impact of prompting LLMs to respond "unsure" when not confident in the answer, which significantly reduces hallucination rates, and explore the phenomenon of "hallucination snowballing," where LMs over-commit to early mistakes, leading to more mistakes. Our findings have implications for the development of more accurate and reliable LLMs and highlight the potential of ChatGPT as a tool for evaluating answer correctness, particularly in mitigating the risks associated with linguistic calibration, counterfactual simulatability, and permutation sensitivity in LLMs.\\
\\\hline
\multirow{2}{*}{ChatGPT Evaluation (OpenAI 2024a)}\\
\\\hline\\
Here is a rating of the abstract based on the provided criteria:\\
1. Ethical Concerns (True/False)\\
False\\
There are no immediate ethical concerns apparent in the abstract. It addresses the use of ChatGPT for evaluating answer correctness and mitigating hallucination, which seems to align with responsible AI practices. However, ethical concerns could arise depending on how the methodology is implemented and its broader implications, but none are highlighted in the abstract.\\
2. Soundness (1-5)\\
4\\
The abstract presents a well-defined and logical approach to evaluating the correctness of LLM outputs, introducing a new metric and methodology. However, the soundness could be rated higher if more information about the evaluation process and its limitations were provided in the abstract.\\
3. Presentation (1-5)\\
4\\
The abstract is well-written, clear, and concise. It effectively communicates the main contributions and findings of the paper. However, it could be slightly more focused by reducing the number of topics covered, such as the detailed exploration of "hallucination snowballing" or "linguistic calibration," which might be better explained in the full paper.\\
4. Contribution (1-5)\\
5\\
The contribution appears significant, as it introduces a novel evaluation methodology, a new metric (ALM), and provides insights into reducing hallucinations in LLMs. The findings regarding high agreement between human judgment and LLM-based metrics are valuable, as well as the investigation into how prompting can mitigate hallucination.\\
5. Overall (1-5)\\
4\\
The abstract suggests a strong and relevant contribution to the field of LLM evaluation. However, a slight reduction in the scope or a more detailed focus on a single key finding could enhance the overall impact.\\
6. Confidence (1-5)\\
4\\
I have a high level of confidence in the proposed methodology and findings based on the abstract. However, my confidence would be higher if there were more details about the experimental setup, sample size, and statistical validation of the results.\\
\\\hline
\end{tabular}
}
\end{table}

\vspace{0pt}
\begin{table}[H]
\label{tab:my-table-18}
  \resizebox{\linewidth}{!}{
\begin{tabular}{|p{10cm}|}
\hline
\\
7. Decision (Accept/Reject)\\
Given the information provided in the abstract and the ratings across various criteria, I would accept the paper for development into an IEEE submission.\\
The work addresses a critical and timely issue in AI—hallucination in large language models—and offers a novel approach to evaluating and mitigating this problem. The introduction of a new metric (ALM) and the exploration of concepts like "hallucination snowballing" demonstrate a significant contribution to the field. While there are areas where more detail could enhance the clarity and soundness, the overall originality, significance, and quality of the work make it a valuable contribution worth pursuing.\\
\\\hline
\end{tabular}
}
Note: The judging criteria were taken from \cite{21}.
\end{table}

\section{Appendix E}

\raggedbottom
\begin{table}[H]
\label{tab:my-table-19}
  \resizebox{\linewidth}{!}{
\begin{tabular}{|p{\linewidth}|}
\hline
\multirow{2}{*}{Sample Topic Tree}\\
\\\hline
\\
Optimization\\\\

1.	https://raw.githubusercontent.com/mlresearch\\/v235/main/assets/sohrabi24a/sohrabi24a.pdf\\
2.	https://openreview.net//pdf?id=iKarSI2a73\\
3.	https://aclanthology.org/2024.emnlp-main.57.pdf\\
4.	https://aclanthology.org/2024.emnlp-main.27.pdf\\
5.	https://openreview.net//pdf?id=VNjJAWjuEU\\
6.	https://raw.githubusercontent.com\\/mlresearch/v235/main/assets/garcin24a/garcin24a.pdf\\\\

Language models\\\\

1.	https://aclanthology.org/2024.emnlp-main.90.pdf\\
2.	https://aclanthology.org/2024.emnlp-main.53.pdf\\
3.	https://aclanthology.org/2024.acl-long.20.pdf\\
4.	https://aclanthology.org/2024.emnlp-main.83.pdf\\
5.	https://aclanthology.org/2024.emnlp-main.37.pdf\\\\

Chain-of-thought reasoning\\\\

1.	https://aclanthology.org/2024.emnlp-main.20.pdf\\
2.	https://aclanthology.org/2024.acl-long.65.pdf\\
3.	https://openreview.net//pdf?id=UdByCgCNdr\\\\

Prompt engineering\\\\

1.	https://aclanthology.org/2024.acl-long.51.pdf\\
2.	https://aclanthology.org/2024.acl-long.65.pdf\\
3.	https://aclanthology.org/2024.acl-long.92.pdf\\
4.	https://aclanthology.org/2024.acl-long.91.pdf\\
5.	https://aclanthology.org/2024.emnlp-main.64.pdf\\
6.	https://aclanthology.org/2024.acl-long.40.pdf\\\\

\\\hline
\end{tabular}
}
\end{table}
\clearpage

\begin{table}
\resizebox{\linewidth}{!}{\begin{tabular}{|p{10cm}|}
\hline\\

Adversarial attacks\\\\

1.	https://openreview.net//pdf?id=OQQoD8Vc3B\\
2.	https://raw.githubusercontent.com/mlresearch\\/v235/main/assets/wang24cn/wang24cn.pdf\\
3.	https://raw.githubusercontent.com/mlresearch\\/v235/main/assets/hu24c/hu24c.pdf\\
4.	https://raw.githubusercontent.com/mlresearch\\/v235/main/assets/bailey24a/bailey24a.pdf\\\\

Benchmarking\\\\

1.	https://aclanthology.org/2024.emnlp-main.27.pdf\\
2.	https://aclanthology.org/2024.emnlp-main.19.pdf\\
3.	https://aclanthology.org/2024.acl-long.83.pdf\\
4.	https://openreview.net//pdf?id=jze2r6RDFz\\
5.	https://aclanthology.org/2024.acl-long.40.pdf\\\\

Cognitive psychology\\\\

1.	https://openreview.net//pdf?id=UdByCgCNdr\\
2.	https://aclanthology.org/2024.emnlp-main.37.pdf\\
3.	https://aclanthology.org/2024.acl-long.93.pdf\\
4.	https://aclanthology.org/2024.acl-long.82.pdf\\
5. https://raw.githubusercontent.com/mlresearch/v235/main/assets/coda-forno24a/coda-forno24a.pdf\\\\

Natural language processing\\\\

1.	https://aclanthology.org/2024.acl-long.57.pdf\\
2.	https://aclanthology.org/2024.acl-long.32.pdf\\
3.	https://aclanthology.org/2024.acl-long.55.pdf\\
4.	https://aclanthology.org/2024.acl-long.22.pdf\\
5.	https://aclanthology.org/2024.acl-long.99.pdf\\\\

Scientific innovation\\\\

1.	https://aclanthology.org/2024.acl-long.18.pdf\\
2.	https://aclanthology.org/2024.emnlp-main.23.pdf\\\\

AI assistants\\\\

1.	https://aclanthology.org/2024.acl-long.61.pdf\\
2.	https://aclanthology.org/2024.emnlp-main.38.pdf\\
3.	https://openreview.net//pdf?id=GEcwtMk1uA\\
4.	https://aclanthology.org/2024.acl-long.40.pdf\\
5.	https://aclanthology.org/2024.emnlp-main.82.pdf\\
\\
\hline
\end{tabular}
}
\end{table}

\section{Appendix F}
\textbf{Context: }From the first abstract, InnoGraph uses a hybrid retrieval-generative model with a dynamic graph. It has a retrieval module and a generative model, and it uses contrastive learning to ensure novelty. They evaluated it with domain experts.
\begin{table}[H]
    \centering
    \resizebox{\linewidth}{!}{
    \begin{tabular}{|p{10cm}|}
    \hline\\
    Sample Experimental Prodcedure\\\\
    \hline\\
        1. **Gather Background Context**: Collect comprehensive background information including problem descriptions, motivations, experimental settings, and constraints.

        2. **Retrieve Relevant Inspirations**: Use a retrieval module to dynamically fetch related past scientific papers and concepts to provide a foundation for generating new ideas.
        
        3. **Generate Initial Ideas**: Employ a generative model to produce initial scientific ideas based on the retrieved inspirations.
        
        4. **Iteratively Improve for Novelty**: Implement an iterative process where each generated idea is compared against existing literature to enhance its novelty. This may involve contrastive learning mechanisms to avoid over-reliance on existing work.
        
        5. **Evaluate with Human Experts**: Conduct evaluations using domain experts to assess the generated ideas on metrics such as relevance, utility, novelty, and technical depth, ensuring the ideas are both innovative and sound.\\\\
        \hline
    \end{tabular}
    }
\end{table}

\end{document}